\documentclass[10pt,twocolumn,letterpaper]{article}

\usepackage[pagenumbers]{cvpr} 
\usepackage{xspace}
\usepackage{multirow}
\usepackage{natbib}
\usepackage[accsupp]{axessibility}

\usepackage[dvipsnames]{xcolor}

\definecolor{cvprblue}{rgb}{0.21,0.49,0.74}
\usepackage[pagebackref,breaklinks,colorlinks,citecolor=cvprblue]{hyperref}

\def\paperID{6147} 
\def\confName{CVPR}
\def\confYear{2025}

\newcommand{\papername}{HyperLoRA\xspace}

\title{\papername: Parameter-Efficient Adaptive Generation for Portrait Synthesis}

\author{
    Mengtian Li\textsuperscript{*} \qquad
    Jinshu Chen\textsuperscript{*} \qquad
    Wanquan Feng\textsuperscript{*$\dagger$} \qquad
    Bingchuan Li \qquad
    Fei Dai \\
    Songtao Zhao \qquad
    Qian He \\[5pt]
    Intelligent Creation, ByteDance
}

\newcommand\nonumfootnote[1]{%
\begingroup%
    \renewcommand\thefootnote{}\footnote{\hspace{-3.7pt}#1}%
    \addtocounter{footnote}{-1}%
\endgroup%
}

\begin{document}

\twocolumn[{%
\renewcommand\twocolumn[1][]{#1}%
\maketitle
\centering
\includegraphics[width=1.95\columnwidth]{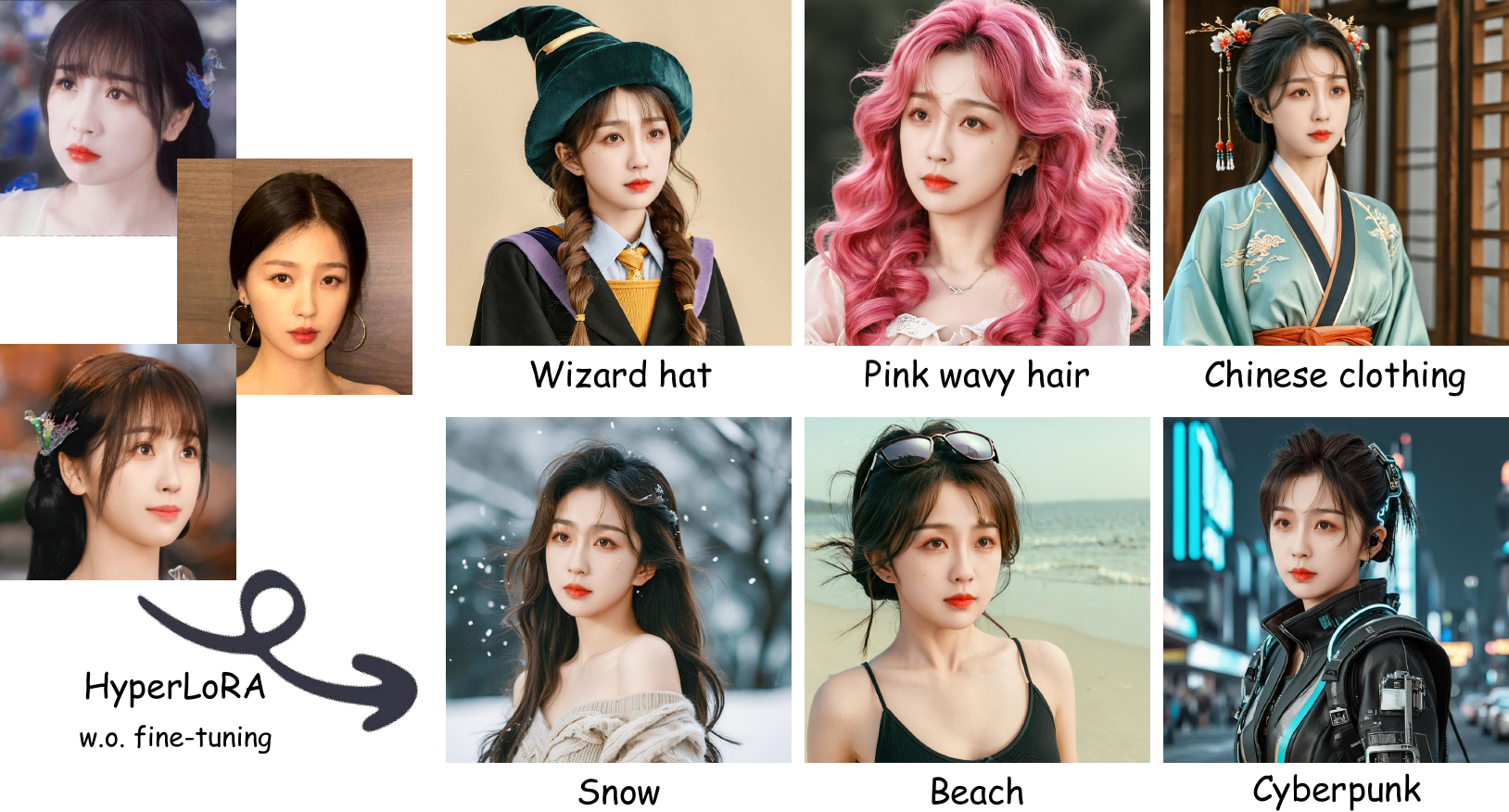}
\captionof{figure}{We propose HyperLoRA, a parameter-efficient adaptive method for portrait synthesis. Given an input face image, HyperLoRA generates personalized LoRA weights without online fine-tuning. Due to the natural interpolability of LoRA, it is easy to support multiple inputs by simple averaging. Leveraging the generated LoRA, we can create personalized portraits with high photorealism and fidelity.}
\label{fig:teaser}
\hspace*{\fill}
}]

\begin{abstract}
\nonumfootnote{* Equal contribution \qquad $\dagger$ Corresponding author}
Personalized portrait synthesis, essential in domains like social entertainment, has recently made significant progress. 
Person-wise fine-tuning based methods, such as LoRA and DreamBooth, can produce photorealistic outputs but need training on individual samples, consuming time and resources and posing an unstable risk. 
Adapter based techniques such as IP-Adapter freeze the foundational model parameters and employ a plug-in architecture to enable zero-shot inference, but they often exhibit a lack of naturalness and authenticity, which are not to be overlooked in portrait synthesis tasks.
In this paper, we introduce a parameter-efficient adaptive generation method, namely \textbf{\papername}, that uses an adaptive plug-in network to generate LoRA weights, merging the superior performance of LoRA with the zero-shot capability of adapter scheme.
Through our carefully designed network structure and training strategy, we achieve zero-shot personalized portrait generation (supporting both single and multiple image inputs) with high photorealism, fidelity, and editability.
\end{abstract}

\vspace{-15pt}
\section{Introduction}
\label{sec:introduction}

The personalized portrait synthesis task~\cite{ye2023ipadapter,wang2024instantid,he2024imagine} aims to generate images sharing the same identity with what is contained in the given single or multiple portrait images.
By feeding the identity information into specific generative model, the task enables the users to create diverse, high-quality and creative personalized portraits, thus holding substantial importance across various domains, such as entertainment and social media.

One key challenge of the personalized portrait synthesis task is the \textbf{fidelity}: the expected generated images should retain the given identity to the greatest extent and avoid irrelevant outputs. Another challenge lies in \textbf{editability}: while preserving the given identity, the portrait synthesis model should maintain the abilities of flexibly editing the other contents of the generated images, for example, the composition of the image, the positioning and the clothing of the character. Additionally, the inference \textbf{speed} is crucial, as fast inference is friendly to not only the computational costs but also the user experience.

Considering the challenges mentioned above, most of the current solutions have different shortcomings. Existing personalized image synthesis methods can generally be divided into tuning-based and tuning-free approaches. 
Tuning-based methods, such as LoRA (Low-Rank Adaption)~\cite{lora} and DreamBooth~\cite{ruiz2022dreambooth}, update model parameters based on specific user-provided data, which can accurately comprehend and preserve the input identity information. Theoretically, these methods are mainly designed for few-shot cases, and customize different models for each groups of provided data with different identities. Thus, disadvantages like high computing costs, unstable training, and lengthy training processes make such methods bloated when dealing with various provided identities.

Tuning-free methods, such as IP-Adapter~\cite{ye2023ipadapter}, typically utilize an additional plug-in structure to encode user identity information. By feeding such identity-related tokens into the frozen foundational model, these methods make influences on the inference process, and ultimately yielding results that retain the given identities. Such additional adapter struture is trained with large portrait datasets in advance, and freezed when inference. Therefore, these methods get rid of the online training, and are convenient for the actual uses. However, for these methods, the interaction between encoded identity information and the foundational model relies on a group of fixed extra cross-attention modules, which might degrade the quality of generated images, leading to unnatural and non-photorealistic facial features, and fail to recover fine-grained details. In summary, the previous methods fail to simultaneously strike a balance between the fidelity, the editability, and the speed.

In this paper, we introduce a parameter-efficient adaptive generation method, namely \textbf{\papername}, which achieves great identity fidelity, high editability and fast inference speed. \papername uses an adaptive plug-in network to encode identity images, which follows the design of tuning-free methods like IP-Adapter. However, unlike such methods which utilize the encoded tokens for the attention computation, we directly generate the LoRA weights with the plug-in network. This strategy not only maintains the zero-shot capability of the adapter strategy, but also inherits the representaion and high-quality generation effect of the LoRA strategy which adjusts the global parameters without real fine-tuning process.

Considering the substantial number of parameters in LoRA, to reduce the training difficulty, we initially project the LoRA parameters onto a linear LoRA parameter space, rather than directly training in the original parameter space. In this way, the generated LoRA weights can be represented as a linear combination of a trainable basis. We employ a perceiver resampler (as in IP-Adapter~\cite{ye2023ipadapter}) to convert the input identity images into combination coefficients for the basis. To encode the input images, we utilize a CLIP ViT~\cite{clip} and an AntelopeV2 encoder~\cite{antelopev2} to extract pixel-level and identity-sensitive features. These feature extractors mentioned above stay frozen during all the training process. To avoid overfitting, we decompose LoRA at the parameter level into an ID-LoRA part and a Base-LoRA part, thereby the ID and ID-irreverent information are fitted into different parts correspondingly. The whole training pipeline runs in a multi-stage way: only the Base-LoRA part is trained in the first stage as a warm-up start; and in the second stage, we train the ID-LoRA with CLIP ViT features; in the final stage, we only finetune the attention modules related to ID embeddings. Besides the case where only a single image is given, our proposed method can also handle cases that there are multiple inputs by simply interpolating the predicted LoRA weights.
Our contributions can be summarized as follows:
\begin{itemize}
    \item We propose \papername, a parameter-efficient adaptive method for the personalized portrait synthesis task, merging the superior performance of the LoRA strategy with the zero-shot capability of the adapter strategy.
    \item We construct a low-dimension linear LoRA space, which reduces the training difficulty of the plug-in network while ensuring the identity preservation.
    \item We design a strategy that decomposes LoRA weights at the parameter level into a base part and an ID part, thereby effectively decoupling the identity information restored in the given images and significantly reducing overfitting.
    \item We conduct various experiments to demonstrate the capacities of our framework. Experimental results show that our method achieves both high fidelity and editability.
\end{itemize}
\section{Related Work}
\label{sec:related_work}

\subsection{Text-to-Image Synthesis}
\label{rw:t2i}
Text-to-image (T2I) synthesis has been rapidly developed for several years. The task aims to generate high-quality images which are highly correlated with given texts or instructions. Some early works~\cite{yu2022scaling,kang2023scaling,tao2022df} employ transformer~\cite{vaswani2017attention} or GAN~\cite{goodfellow2014generative} architectures to solve this task, while the performance of the solutions is not that satisfying owing to the limited capacities of their pipelines. Recently, rapid progress has been made by the methods adopting diffusion pipelines~\cite{23_sohl2015deep,24_ho2020denoising,25_song2020score,27_kingma2021variational,30_dhariwal2021diffusion,31_ho2022cascaded}. With a stepwise noising-denoising design, the generated results of diffusion pipelines have far higher qualities than other self-supervised methods, especially on the datasets with huge intra-domain variance \cite{33_imagenet,34_schuhmann2021laion}. In this paper, we follow a diffusion pipeline \cite{02_sd} to build up our pipeline to ensure the qualities and the varieties of the outputs.

\subsection{Tuning-based Personalized Portrait Synthesis}
\label{rw:trainingbased}
Tuning-based methods finetune models based on user-provided data to comprehend user identity information. Dreambooth~\cite{ruiz2022dreambooth} and its derivative versions~\cite{dreambooth++,hyper-dreambooth,pang2024attndreambooth} treat the entire T2I model trainable, and utilize certain special captions to represent the given identities. LoRA~\cite{lora} and its variants~\cite{DBLP:conf/icml/HayouG024,DBLP:conf/iclr/ZhangCBH0CZ23,DBLP:conf/icml/LiuWY0WCC24,lycoris,han2023svdiff,gal2023encoder} represent another category of tuning-based methods. Unlike Dreambooth, LoRA greatly enhances the flexibility of the fine-tuning scheme by creating several trainable low-rank matrices within the T2I model along with much fewer parameters than the whole model. Nevertheless, a common issue with all the aforementioned methods is that they lack the ability to handle zero-shot cases. Each fine-tuned model matches only a single customized task, thus the fine-tuning incurs a high burden in terms of time, computational resources, and storage resources. In this work, our proposed \papername does not rely on online training of user-provided data, thereby circumventing the resource consumption problem brought about by the training process.

\subsection{Tuning-free Personalized Portrait Synthesis}
\label{rw:trainingfree}
Tuning-free methods, such as~\cite{wang2024instantid,ye2023ipadapter,li2024photomaker,t2i-adapter,ma2024subject}, pre-design a plug-in structure containing few parameters that can be inserted into the T2I base model. After pre-training and freezing such parameters on a large-scale dataset, tuning-free methods utilize such insertable structure to extract the identity information features from the given inputs, and inject them into the base model during inference. The online tuning is not necessary for these methods, thereby they can save resource costs and time. Concurrently, the design of the detachable parameters is also rather flexible. 
However, these methods merely learn the mapping relationship between the input image and the encoded tokens, which are utilized by a group of additional cross-attention modules, and the outputs are simply blended with the original ones. Our proposed HyperLoRA learns the mapping relationship between the input image and the network parameters, which is more essential. The implicit information contained in the network parameters is far more abundant and friendly to the base model than that in tokens. Compared with the methods mentioned above, our method has better adaptability, higher nonlinearity, and better fidelity.

\section{Method}
\label{sec:method}

\begin{figure*}[t]
	\centering
	\includegraphics[width=2\columnwidth]{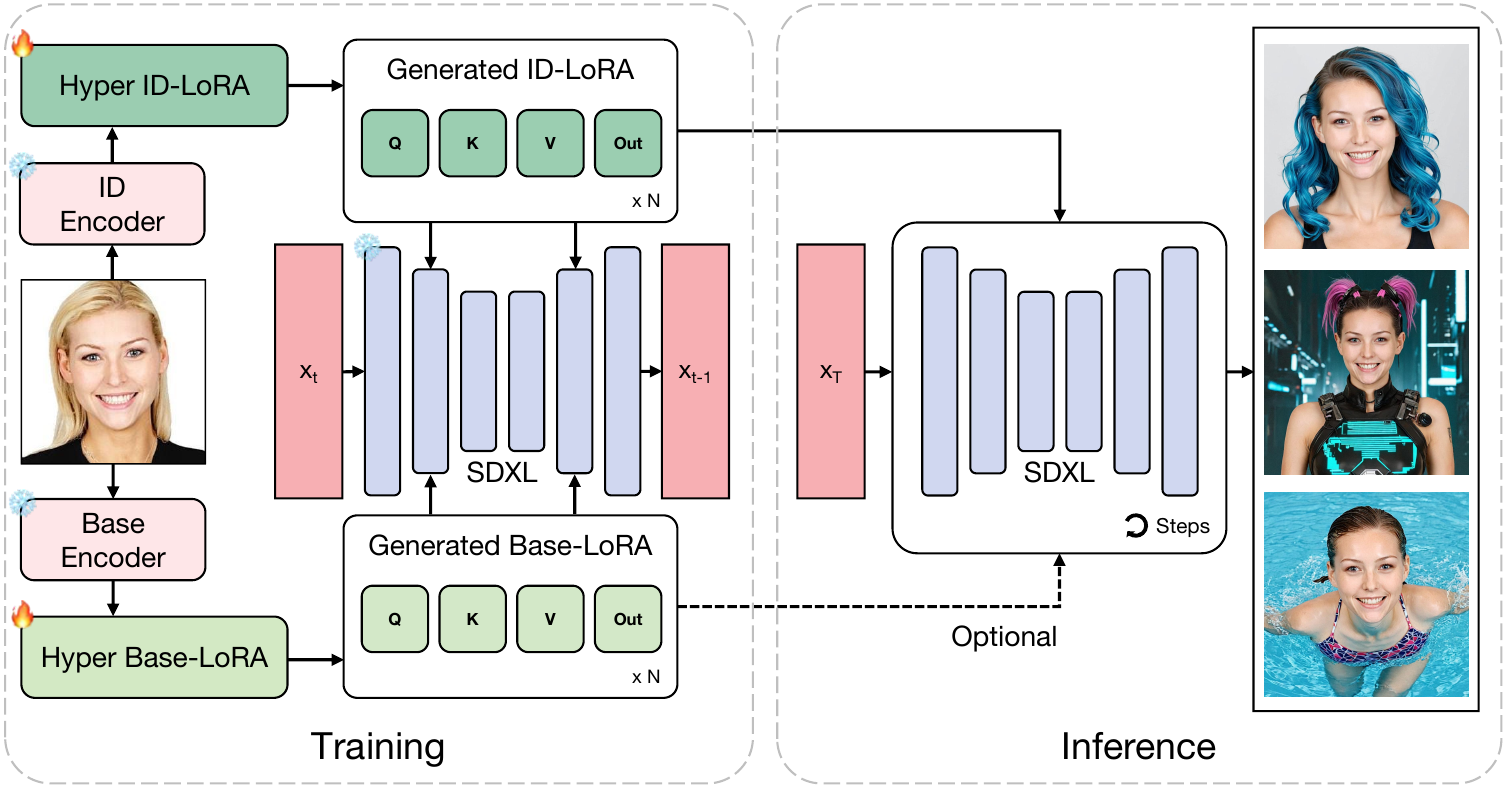}
	\caption{Overview of HyperLoRA. We explicitly decompose the HyperLoRA into a Hyper ID-LoRA and a Hyper Base-LoRA. The former is designed to learn ID information while the latter is expected to fit others, e.g. background and clothing. Such a design helps to prevent irreverent features leaking to ID-LoRA. During the training, we fix the weights of the pretrained SDXL base model and encoders, allowing only HyperLoRA modules updated by Backpropagation. At the inference stage, the Hyper ID-LoRA integrated into SDXL generates personalized images while the Hyper Base-LoRA is optional.}
	\label{fig:overview}
\end{figure*}

In this section, we introduce the details of our method, overview of which is illustrated in Fig.~\ref{fig:overview}. In Sec.~\ref{sec:lora_space}, we introduce the low-dimensional linear LoRA space. In Sec.~\ref{sec:network}, we introduce the network structure. In Sec.~\ref{sec:training}, we introduce the designed multi-stage training process.

\subsection{Low-Dimensional Linear LoRA Space}
\label{sec:lora_space}
The number of parameters in LoRA is typically large, which makes it difficult to generate the LoRA weights directly. Recognizing the interpolative nature of LoRA, we are inspired to exploit its linearity, and map the LoRA parameters onto a low-dimensional linear space. Note that our discussion below and in the rest of the paper is based on the implementation from ~\cite{Ryu_Low-rank_adaptation_for}, which is the most widely acknowledged way. Each layer of LoRA includes $8$ matrices, which consist of the down and up matrices for query, key, value, and output. We project each matrix onto a $K$-dimensional basis. Mark the number of LoRA layers as $N$, then the overall degrees of freedom for the entire LoRA can be calculated as $N\times8K$, which is significantly smaller than the origin parameter number in most cases. In our paper, we use $K=128$, and the number of parameters after projection becomes approximately only $1.2\%$ ($0.14$M vs. $11.6$M when the LoRA rank is $8$) of the original one. To decouple the given specific identity information from the irrelevant features, we design to split a LoRA matrix into a base part and an ID part. We denote the basis of the ID part as $\{\mathbf{M}_{id}^{k}\}_{k=1}^{K}$, the combination coefficients as $\{\alpha_{k}\}_{k=1}^{K}$. Further, we denote the ID part as:
\begin{equation}
\begin{split}
\mathbf{M}_{id} = \sum\limits_{k=1}^{K} \alpha_{k} \cdot \mathbf{M}_{id}^{k}.
\label{eq:idlora_formula}
\end{split}
\end{equation}

Similarly, the base part, denoted as $\mathbf{M}_{base}$, follows the same format as $\mathbf{M}_{id}$:
\begin{equation}
\begin{split}
\mathbf{M}_{base} = \sum\limits_{k=1}^{K} \beta_{k} \cdot \mathbf{M}_{base}^{k}.
\label{eq:baselora_formula}
\end{split}
\end{equation}

Although the mathematical format of Base-LoRA is similar to ID-LoRA, it possesses a different significance. Specifically, we configure $\mathbf{M}_{base}$ to contain identity-irrelevant contents across all the training samples, such as the background and clothing, essentially capturing information from the input image that should not be included in $\mathbf{M}_{id}$. Explicitly modeling Base-LoRA can help ID-LoRA learn more essential identity-related information and ignore the irrelevant rest. In summary, we represent a \papername parameter matrix $\mathbf{M}$ as:
\begin{equation}
\begin{split}
\mathbf{M} = \mathbf{M}_{base} + \mathbf{M}_{id} = \sum\limits_{k=1}^{K} \beta_{k} \cdot \mathbf{M}_{base}^{k} + \sum\limits_{k=1}^{K} \alpha_{k} \cdot \mathbf{M}_{id}^{k}.
\label{eq:lora_formula}
\end{split}
\end{equation}

\begin{figure}[t]
	\centering
	\includegraphics[width=1\columnwidth]{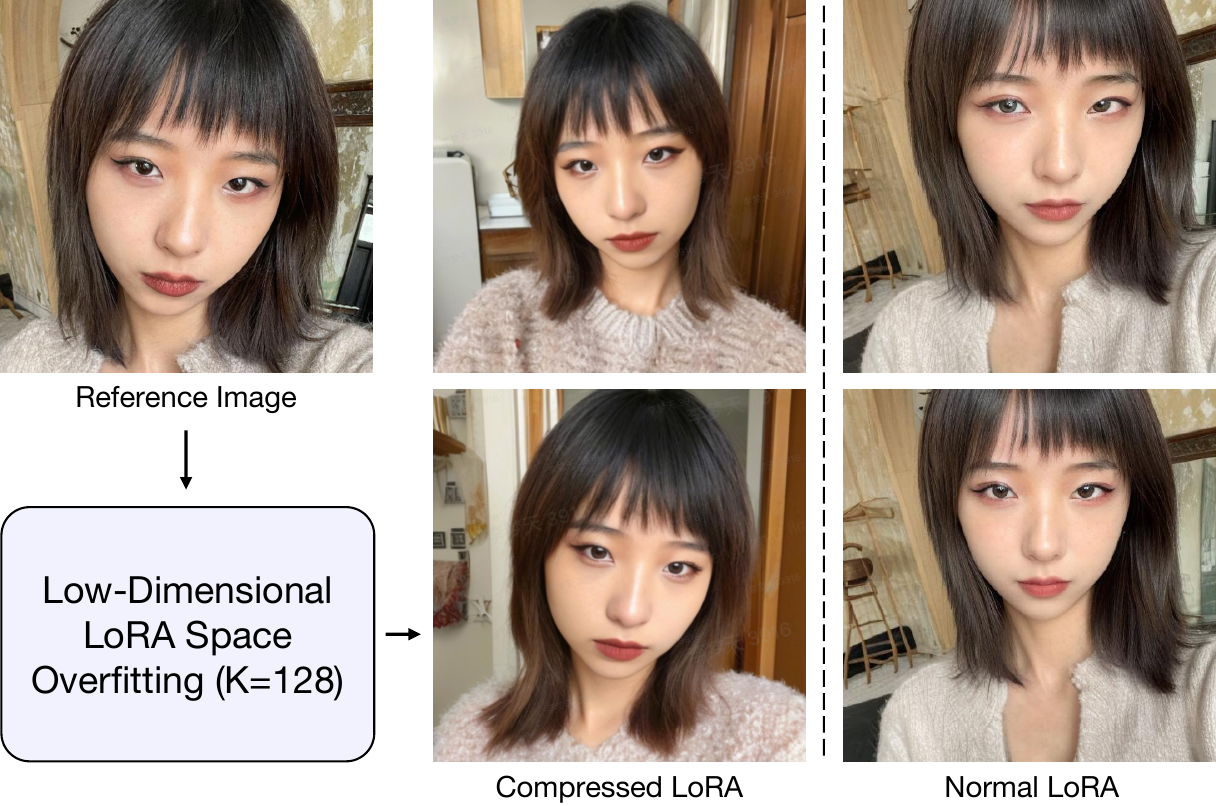}
	\caption{The identity reconstruction ability on the low-dimensional linear LoRA space. In this example, we project LoRA parameters onto 128-dim basis and train on a face dataset (about 400K samples). Compared to normal LoRA, our compressed LoRA can also maintain the identity of the reference image well.}
	\label{fig:space}
\end{figure}

It is noteworthy that although the degrees of freedom for the LoRA parameters are significantly compressed, they still retain the capacity to represent identity information, as Fig.~\ref{fig:space} shows. With our designed projection method and the setting of $K=128$, the generated results in the illustration of overfitting a face dataset suggest that, it is still capable of reconstructing the identity of the reference image for the compressed LoRA module without obvious degradation.

\subsection{Network Structure}
\label{sec:network}

\begin{figure}[t]
    \centering
    \includegraphics[width=1\columnwidth]{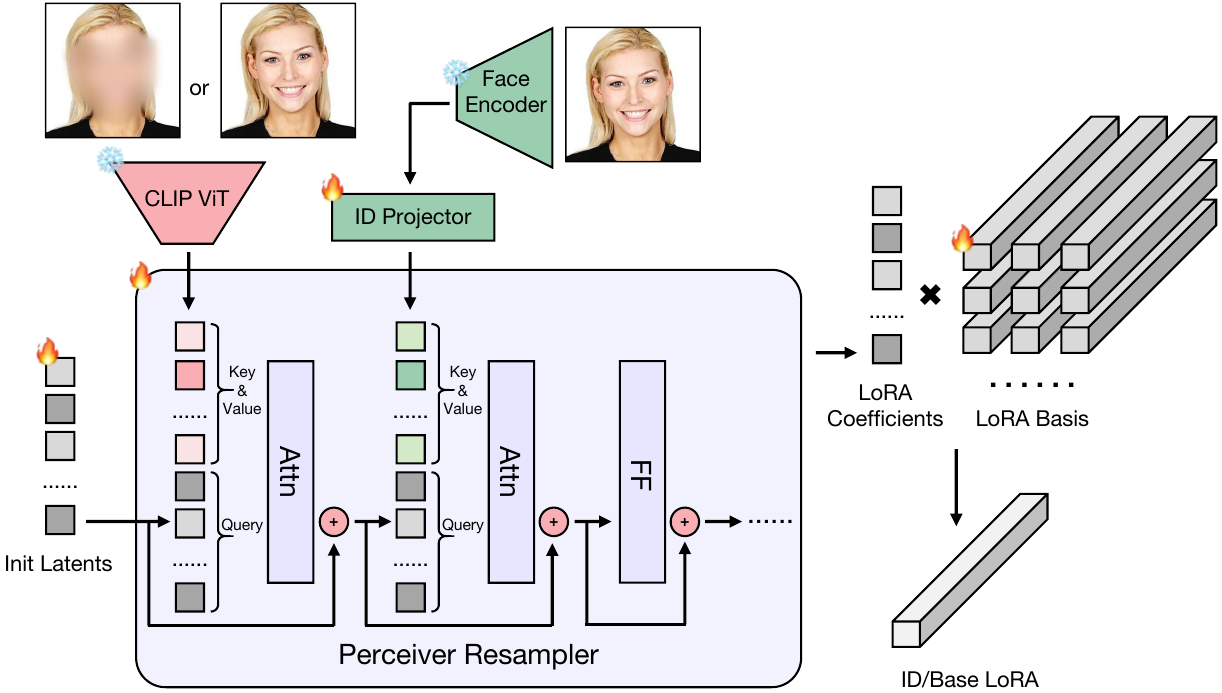}
    \caption{Network structure of HyperLoRA. We apply a perceiver resampler to convert the image features into a group of LoRA coefficients, thereby generating the whole LoRA by multiplied with LoRA basis. Two independent perceiver resamplers are instantiated for Hyper Base and ID LoRAs. Note that the second attention block (interacting with green tokens from ID Projector) is absent in Hyper Base-LoRA.}
    \label{fig:network}
\end{figure}

In this section, we introduce our network structure, which is shown in Fig.~\ref{fig:network}. The whole network can be divided into several main components: the image encoder, the preceiver resampler and the LoRA combination. 

\noindent \textbf{Image Encoder.} The input image is passed into two image encoders, including a CLIP~\cite{clip} ViT and an AntelopeV2~\cite{antelopev2} encoder. The features extracted by the former are dense and contain strong structural information, while the features extracted by the latter are more abstract identity information. To explicitly distinguish the training processes of the base part and the id part, we deal with the input image of the CLIP in two different ways: for the training of $\mathbf{M}_{base}$, we first crop the face region of an original image, and inside the cropped image, we blur the facial features and contour, preventing the base part reconstructing the specific face; for the training of $\mathbf{M}_{id}$, we directly feed the cropped face image into CLIP ViT. Meanwhile, it also serves as the input of AntelopeV2 encoder to extract ID embedding. We utilize a simple ID Projector, consisting of a Linear layer followed by a Reshape and a LayerNorm~\cite{ba2016layer}, to transform the ID embedding into some ID tokens. The details of the approaches mentioned above can be found in Fig.~\ref{fig:network}.

\noindent \textbf{Perceiver Resampler.} The perceiver resampler~\cite{alayrac2022flamingo,ye2023ipadapter} employs a transformer backbone and incorporates image tokens into the computation through an attention mechanism. Currently, we apply a 4-layer perceiver resampler, each layer of which consists of two attention modules interacting with the tokens from CLIP ViT and ID Projector. Note that we skip the second attention module when training $\mathbf{M}_{base}$. We obtain the output of the perveiver resampler, the combination coefficients $\{\alpha_{k}\}_{k=1}^{K}$ and $\{\beta_{k}\}_{k=1}^{K}$, which would be used to compute the output LoRA weights subsequently.

\noindent \textbf{LoRA Combination.} Both the Base-LoRA basis $\{\mathbf{M}_{base}^{k}\}_{k=1}^{K}$ and the ID-LoRA basis $\{\mathbf{M}_{id}^{k}\}_{k=1}^{K}$ are trainable variables. They are optimized during the training phase and fixed as constants during inference. Once the combination coefficients $\{\alpha_{k}\}_{k=1}^{K}$ and $\{\beta_{k}\}_{k=1}^{K}$ are predicted, the network can compute the entire LoRA by Eq.~\ref{eq:lora_formula}. It is noteworthy that the combination way can be adjusted flexibly to achieve various effects, for example, if the editability is a priority, the weight of $\mathbf{M}_{base}$ in Eq.~\ref{eq:lora_formula} can be reduced to drop the information unrelated to the input identity.

\subsection{Training}
\label{sec:training}
In this section, we introduce our multi-stage training process. Our approach begins with training the Base-LoRA as an initial warm-up. Subsequently, we expand the training to ID-LoRA with CLIP ViT. Finally, we fine-tune the model to enhance its identity preserving capability by using AntelopeV2 face encoder. For all stages, we employ the same loss term as in DDPM~\cite{24_ho2020denoising}.

\noindent \textbf{Base-LoRA Warm-up.} Training the ID-LoRA directly from an open domain base model is quite challenging, as it tends to inadvertently incorporate extraneous details such as composition, background, and attire. Therefore, as described in Sec.\ref{sec:lora_space}, we train non-facial information into the Base-LoRA, which serves as a warm-up in our training process. In this stage, we blur the facial information of the original input image before passing it through the CLIP ViT, and bypass the AntelopeV2 face encoder. 

\noindent \textbf{ID-LoRA with CLIP.} After the Base-LoRA warm-up, we add ID-LoRA into the training process. In this stage, we add the AntelopeV2 branch besides the existed CLIP branch, so that the abstract identity information can be efficiently extracted. We observe that training with CLIP features converges quickly but tends to overfit structural details, whereas training with ID embedding converges more slowly. Consequently, in this stage we stop training with CLIP features as long as it reaches the approximate convergence. To stabilize the training state in the early iterations, we initialize the output layers of the attention modules that are fused with ID embeddings to zero, allowing only the CLIP features to guide the early generation. To enable ID-LoRA to effectively represent the identity in the presence of trigger words (e.g., 3 rare words \texttt{fcsks fxhks fhyks} in this paper), while ensuring that the results remain unaffected in their absence, each iteration of this stage is set to one of the following situations: $\mathbf{1).}$ The given prompt contains trigger words, both Base-LoRA and ID-LoRA are enabled, and the reconstruction target is the target image; $\mathbf{2).}$ The given prompt does not contain trigger words, only ID-LoRA is enabled, and the resulting noise prediction should align with that of the original foundational model; $\mathbf{3).}$ The given prompt does not contain trigger words, both Base-LoRA and ID-LoRA are enabled, and the resulting noise prediction should be consistent with that when only Base-LoRA is enabled. This approach minimizes disruption to the base model parameters and further concentrates the parameters on those related to identity information.

\noindent \textbf{ID-Embedding Fine-tune.} After the previous training stages achieve rough convergence, we fix the weights of the CLIP-related attention modules, and continue to fine-tune the attention modules related to ID embedding (the second attention in Fig.~\ref{fig:network}). The training loss of this stage is exactly the same as in the second stage.
\section{Experiments}
\label{sec:experiments}

\subsection{Implementation Details}
\label{sec:experiments:implementation}

To train our HyperLoRA model, we collect a subset from LAION-2B~\cite{schuhmann2022laion} which comprises $4.4$ million portrait images. In detail, InsighFace~\cite{insightface} is leveraged to detect faces and we filter out those non-single portrait images. Meanwhile, we generate a caption for each image in our dataset using BLIP-2~\cite{li2023blip} to enable Text-to-Image training. During the training, we randomly crop images around the face for data augmentation and focus more on the face region.
Considering memory consumption, we currently set the LoRA rank of ID-LoRA and Base-LoRA to $8$ and $4$ respectively, which costs about $60$G memory when batch size is $1$. We further apply the 8-step gradient accumulation to simulate a larger batch size. 
We train the model based on SDXL-Base-1.0~\cite{podell2023sdxl} with AdamW~\cite{loshchilov2017decoupled} optimizer and learning rate $0.00001$. In this 3-stage training process, we optimize the Base-LoRA, ID-LoRA (CLIP) and ID-LoRA (ID embedding) for $20$K, $15$K and $55$K iterations in turn, and set the probability of three situations to $0.9$, $0.05$ and $0.05$ for ID-LoRA training. We implement our method with PyTorch and Diffusers. The total training process is performed on 16 NVIDIA A100 GPUs for about 10 days.

\begin{figure}[t]
    \centering
    \includegraphics[width=1\columnwidth]{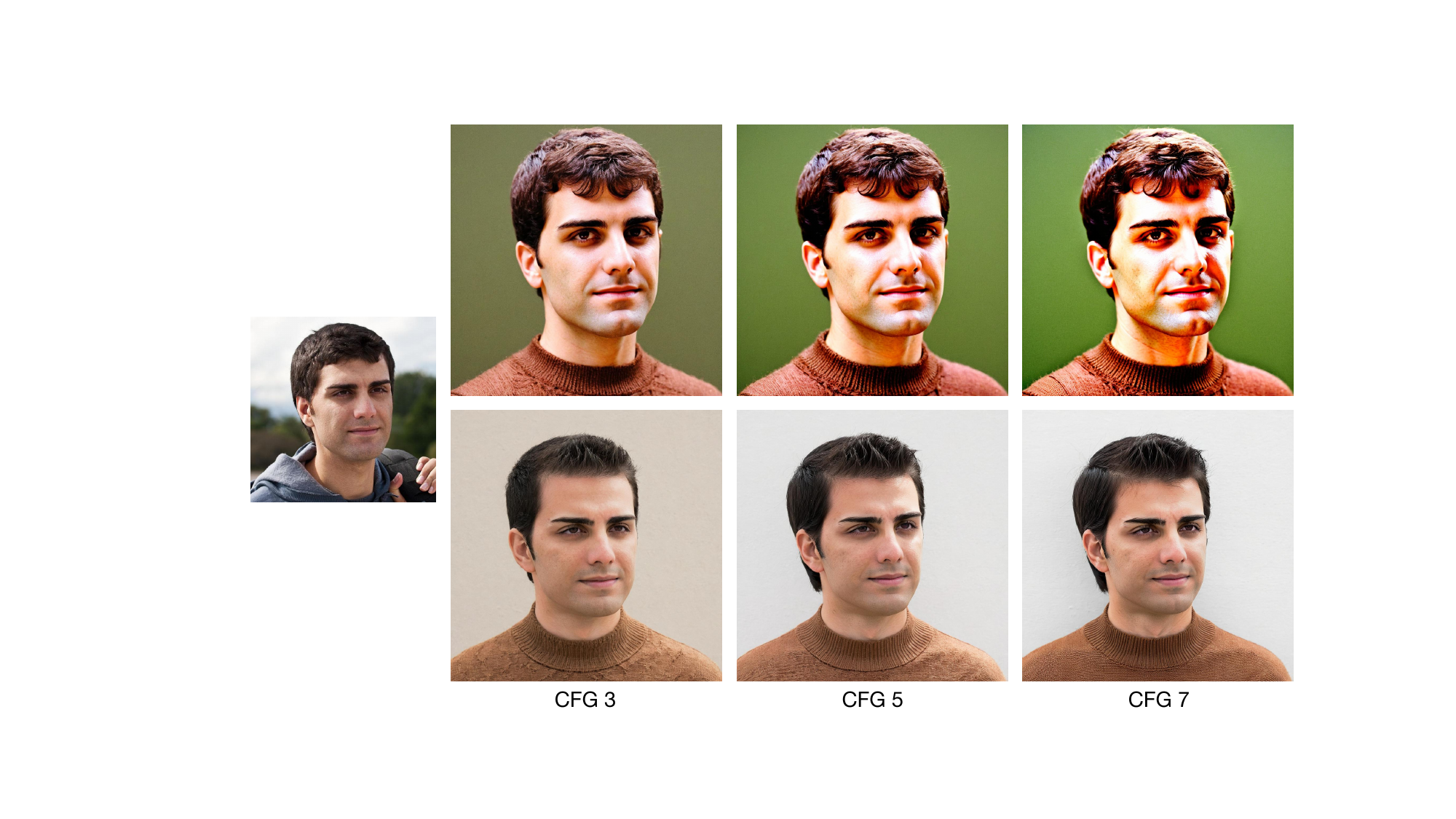}
    \caption{Adapter and LoRA show different tolerances to CFG. The first row: images generated by InstantID where higher CFG leads to oversaturation. The second row: HyperLoRA always yield reasonable portrait images for CFG ranged from $3$ to $7$.}
    \label{fig:cfg}
\end{figure}

\begin{figure}[t]
    \centering
    \includegraphics[width=1\columnwidth]{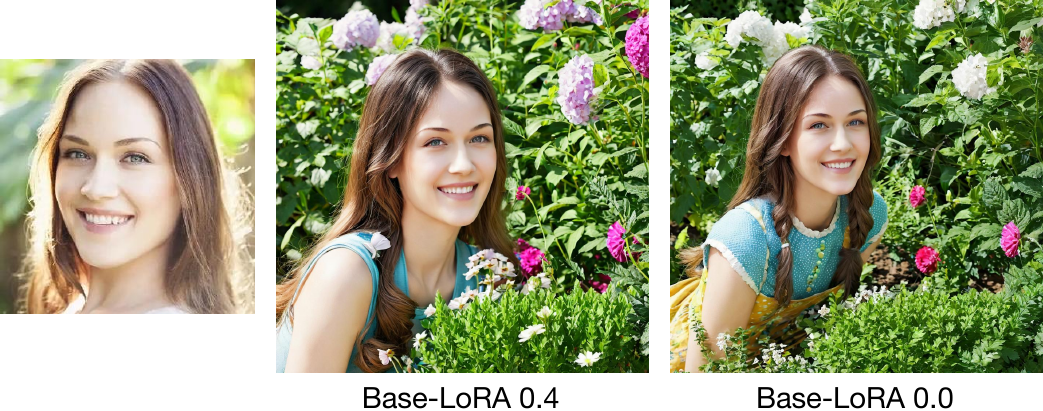}
    \caption{Inference with Base-LoRA improves face fidelity but limits the layout of generated images more similar to the cropped region for training.}
    \label{fig:baselora}
\end{figure}

\subsection{Settings}
\label{sec:experiments:settings}
We build workflows in ComfyUI~\cite{comfyui} to evaluate our HyperLoRA and three SOTA ID preservation methods, i.e. InstantID~\cite{wang2024instantid}, IP-Adapter~\cite{ye2023ipadapter} (specifically, IP-Adapter-FaceID-Portrait) and PuLID~\cite{guo2024pulid}. To ensure a fair comparison, we apply the same seed, prompt and sampler across all methods. Note that we additionally concatenate the lora trigger words as a prefix to the prompt when testing HyperLoRA, and set different Classifier-Free Guidance~\cite{ho2022classifier} (CFG) scales to generate reasonable images. As shown in Fig.~\ref{fig:cfg}, Adapter-based methods usually require a lower CFG scale while HyperLoRA allows a wide range of CFG scales.
Although these methods are trained on SDXL-Base-1.0, they show sufficient generalization to other SDXL models. Thus, we use LEOSAM~\cite{leosams}, an open-source SDXL model for high quality portraits, as the base model for evaluations, which typically improve image aesthetics. In quantitative comparison, we utilize a recent open-source test set, Unsplash-50~\cite{gal2024lcm}, which covers diverse genders, ages and races.

\begin{table}[h]
    \centering
    \caption{Quantitative comparison.}
    \setlength{\tabcolsep}{5pt}
    \begin{tabular}{lcccc}
        \hline
         & \multicolumn{2}{c}{Fidelity} & \multicolumn{2}{c}{Editability} \\
         & CLIP-I$\uparrow$ & ID Sim.$\uparrow$ & CLIP-I$\uparrow$ & CLIP-T$\uparrow$ \\
        \hline
        IP-Adapter & 0.764 & 0.566 & 0.725 & 0.244 \\
        InstantID & 0.734 & \textbf{0.681} & 0.688 & 0.237 \\
        PuLID & 0.771 & 0.613 & \textbf{0.805} & \textbf{0.259} \\
        \hline
        Arc2Face & 0.786 & 0.643 & - & - \\
        \hline
        Ours (Full) & \textbf{0.853} & \underline{0.678} & 0.710 & 0.243 \\
        Ours (ID) & \underline{0.831} & 0.625 & \underline{0.748} & \underline{0.252} \\
        \hline
    \end{tabular}
    \label{tab:quant_comp}
\end{table}

\begin{figure*}[t]
    \centering
    \includegraphics[width=2\columnwidth]{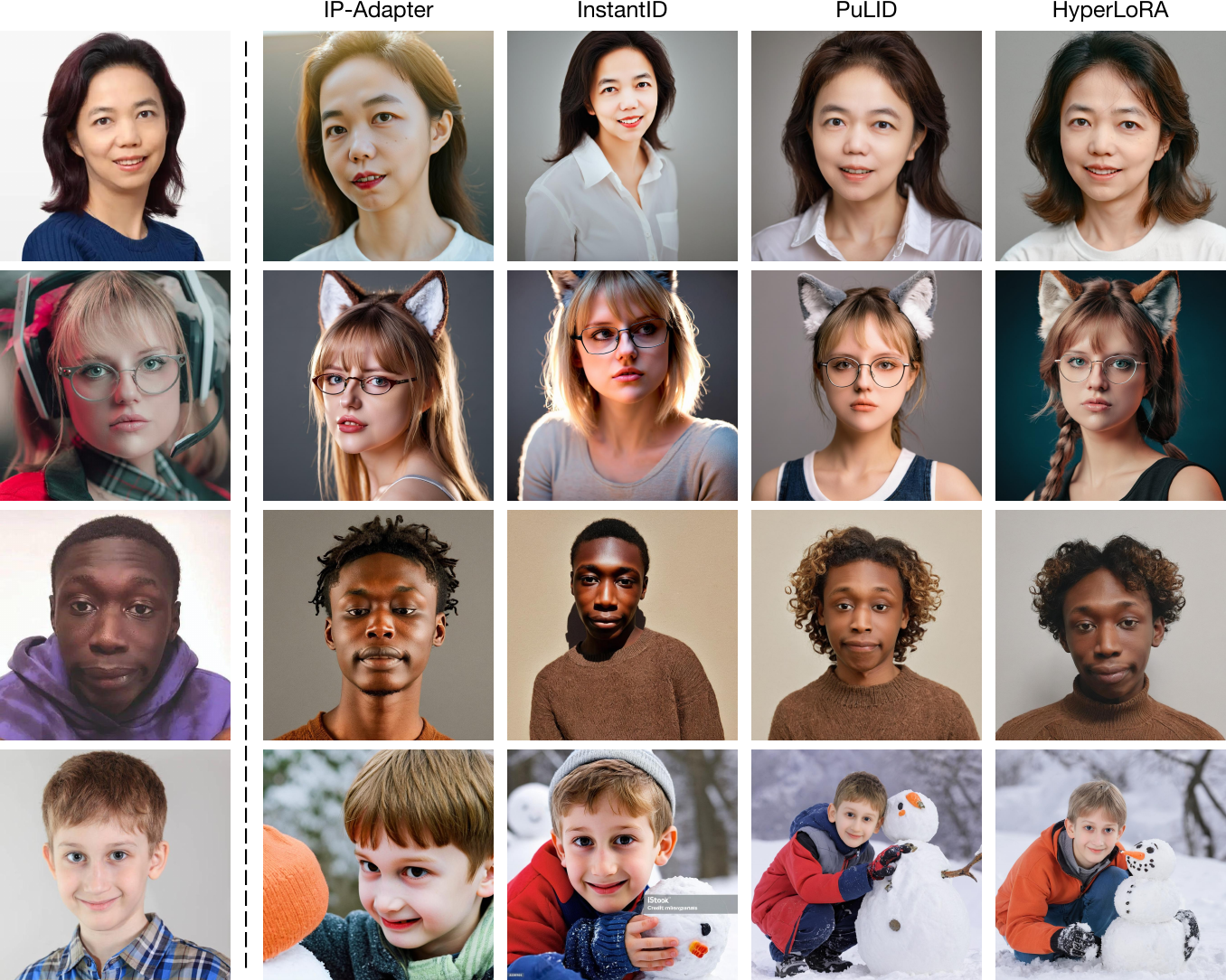}
    \caption{Qualitative comparison. From top to bottom, we generate images with the scene prompts: \textbf{white shirt}, \textbf{wolf ears}, \textbf{wavy hair and brown sweater}, and \textbf{play with snowman}.}
    \label{fig:qual_cmp}
    \vspace{-4pt}
\end{figure*}

\subsection{Comparisons}
\label{sec:experiments:comparisons}

\noindent \textbf{Quantitative Comparison.}
We evaluate fidelity and editability as the quantitative metrics for comparison across these methods. 
For fidelity, we leverage the cosine of CLIP image features (CLIP-I) to measure face fidelity, and the cosine of ID embeddings (ID Sim.) to measure face ID similarity. We generate four images with prompt \texttt{headshot of a girl/boy}, \texttt{high definition}, \texttt{4k} for each sample in our test set using each method, and compute the average of them.
For editability, we utilize CLIP-I to measure the similarity of images generated with and without ID plug-in networks, and the cosine of CLIP image and text features (CLIP-T) to measure image-text similarity. We collect 10 prompts which cover hair, accessory, clothing and background, and evaluate all combinations of prompts and ID images.
The specific CLIP model we used is CLIP-L/14 which is different from the text encoder in SDXL, and CurricularFace~\cite{huang2020curricularface} (not used as ID encoder in all methods) is adopted to extract ID embeddings.
We further incorporate Arc2Face~\cite{papantoniou2024arc2face}, a human face base model, into comparison, which is trained using a fixed caption and a cropped region of faces, thus restricting its editability.
Ours (Full) in Table.~\ref{tab:quant_comp} means inference with both Base-LoRA (weight is set to $0.4$) and ID-LoRA, while Ours (ID) represents only using ID-LoRA.
As shown in Table.~\ref{tab:quant_comp} and Fig.~\ref{fig:baselora}, Base-LoRA helps to improve fidelity but limits editability. Although our design decouples the image features into different LoRAs, it's hard to avoid leaking mutually. Thus, we can adjust the weight of Base-LoRA to adapt to different application scenarios.
Our HyperLoRA (Full and ID) achieve the best and second-best face fidelity while InstantID shows superiority in face ID similarity but lower face fidelity. Both these metrics should be considered together to evaluate fidelity, since the face ID similarity is more abstract and face fidelity reflects more details. This is evidenced in the following visual results in Fig.~\ref{fig:qual_cmp}.
In the aspect of editability, PuLID possesses a clear advantage, the key contribution of which is a well-designed loss function to reduce intrusion into the base model, thereby improving editability. Despite HyperLoRA (ID) inferior to PuLID, it surpasses other methods, meanwhile has good fidelity. It is noteworthy that there is a trade-off between fidelity and editability, and our HypeLoRA reaches a proper balance.

\noindent \textbf{Qualitative Comparison.}
Fig.~\ref{fig:qual_cmp} shows the visual comparisons across IP-Adapter, InstantID, PuLID and our HyperLoRA (only ID-LoRA).
The skin of portraits generated by IP-Adapter and InstantID has apparent AI-generated texture, which is a little oversaturation and far from photorealism. It is a common shortcoming of Adapter-based methods. PuLID improves this problem by weakening the intrusion to base model, outperforming IP-Adapter and InstantID but still suffering from blurring and lack of details. In contrast, LoRA directly modifies the base model weights instead of introducing extra attention modules, usually generating highly detailed and photorealistic images. Our HyperLoRA naturally inherits this advantage from LoRA.
Furthermore, the manner in which LoRA injects ID features exhibits higher nonlinearity. Thus, our method achieves better face fidelity, capturing fine-grained details, such as pupil color.
The layout of generated images of our HyperLoRA and PuLID is more similar compared with other methods, which is because of the higher CLIP-I and CLIP-T of editability. In comparison, IP-Adapter and InstantID have inferior editability, resulting in unreasonable layout (the 4th row of IP-Adapter) and failure to follow the prompt (the 3rd row of InstantID).

\subsection{Ablation Study}
\label{sec:experiments:ablation_study}

\noindent \textbf{No Hyper Base-LoRA.}
\begin{figure}[t]
    \centering
    \includegraphics[width=1\columnwidth]{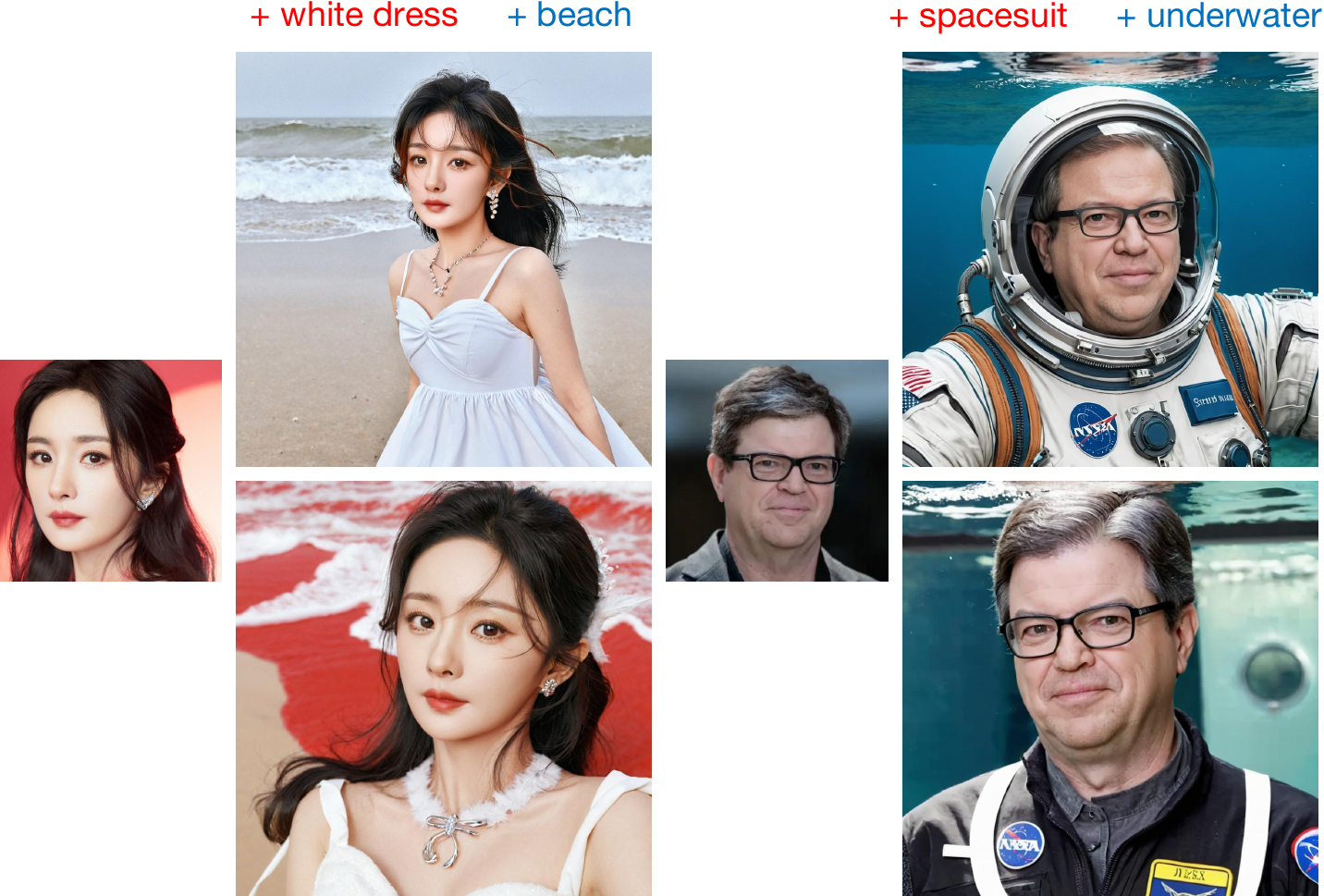}
    \caption{Impact of Hyper Base-LoRA. The first row: the model trained with Hyper Base-LoRA generates correct clothing and background. The second row: the model trained without it takes many irrelevant features into the generated images.}
    \label{fig:ab_hyper_base}
\end{figure}
In Fig.~\ref{fig:ab_hyper_base}, we demonstrate the effectiveness of Hyper Base-LoRA. As aforementioned, our Hyper Base-LoRA helps to decouple ID information from input images. This is evidenced in the second row of Fig.~\ref{fig:ab_hyper_base}, where the model trained without Hyper Base-LoRA fails to properly reflect the prompts \texttt{dress}, \texttt{spacesuit} and \texttt{beach}. As expected, the background and clothing are more editable if trained with Hyper Base-LoRA, since it prevents the ID-irrelevant features leaking to ID-LoRA.

\noindent \textbf{Only CLIP Encoder.}
\begin{figure}[t]
    \centering
    \includegraphics[width=1\columnwidth]{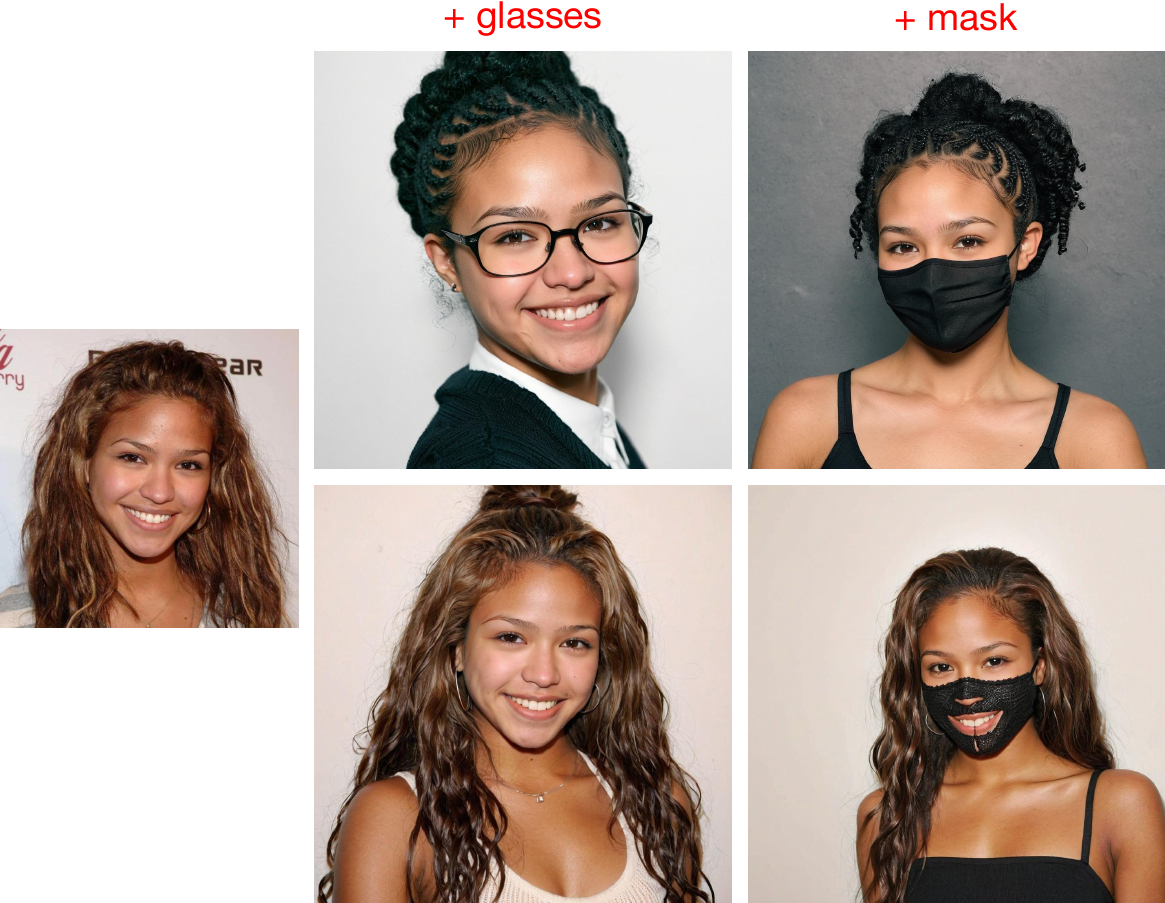}
    \caption{Comparison between different encoders. The first row: our CLIP + ID embeddings training strategy successfuly generates glasses and mask. The second row: training with only CLIP fails to follow the prompts.}
    \label{fig:ab_clip}
\end{figure}
Our multi-stage training strategy is crucial for successfully training ID-LoRA. Especially, we leverage CLIP ViT features only during the early iterations of training and then switch to ID embeddings to learn more details. Fig.~\ref{fig:ab_clip} shows that the problem arises when only adopting CLIP as encoder, where we see that the facial elements are challenging to modify. In contrast, predominantly learning facial features with ID embeddings relaxes the structural constraints on the face.

\noindent \textbf{Multiple Input Images.}
\begin{figure}[t]
    \centering
    \includegraphics[width=1\columnwidth]{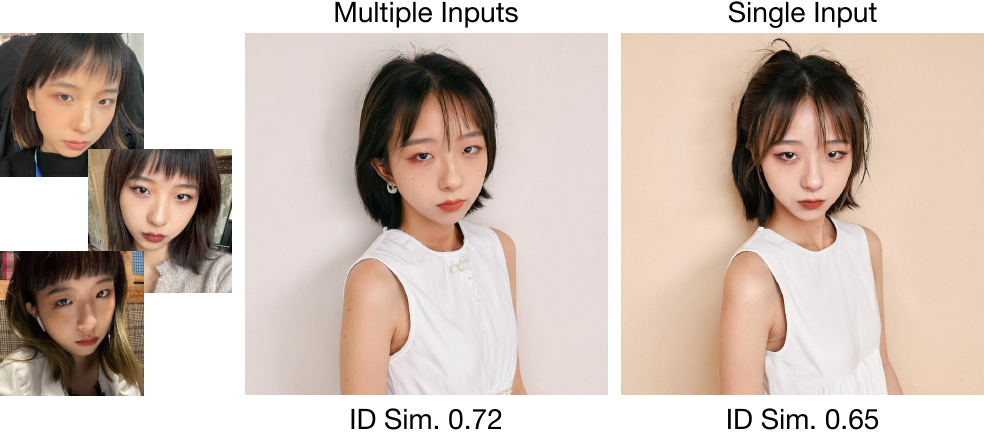}
    \caption{Multiple input images contribute to a higher ID similarity and a more realistic and aesthetic synthesized portrait.}
    \label{fig:ab_multi}
\end{figure}
Thanks to the natural interpolability of LoRA and our low-dimensional linear LoRA space (as discussed in Sec.~\ref{sec:lora_space}), it is straightforward to support portrait synthesis with multiple input images by averaging the LoRA coefficients generated from each single image. Fig.~\ref{fig:ab_multi} illustrates that multiple ID images help HyperLoRA to identify more stable and pure ID features, thereby improving both ID fidelity and image quality.

\section{Conclusion}
\label{sec:conclusion}

We have proposed HyperLoRA, the first zero-shot personalized portrait generation method based on LoRA, which is trained in an end-to-end manner. With the advantage of LoRA, our method modifies the base model weights to inject ID information rather than introducing new cross attention as adapter-based methods do, exhibiting higher nonlinearity, thereby yielding more detailed and realistic portrait images. To make HyperLoRA training feasible, we represent the whole LoRA parameters with a linear combination of a group of LoRA basis, which reduces the number of HyperLoRA parameters magnificently. We further decompose the LoRA into two parts explicitly. With our well-designed training strategy, the ID part tends to learn facial features while other aspects, like background and clothing, are fitted into the base part, thereby reducing the risk of overfitting and improving editability.
It's recognized that only a rank $8$ HyperLoRA is allowed to train when considering the memory limitation of current GPU. Nevertheless, rank $8$ is still significantly smaller than what is frequently used in normal LoRA. An advanced LoRA parameters compression method, such as LoRA autoencoder, may enable larger rank HypeLoRA training, unlocking more powerful capability.

\newpage
\maketitlesupplementary

\appendix
\section{Appendix}
\label{sec:appendix}

\subsection{Inference Speed}
We evaluate the inference speed of IP-Adapter~\cite{ye2023ipadapter}, InstantID~\cite{wang2024instantid}, PuLID~\cite{guo2024pulid} and our HyperLoRA on a single NVIDIA V100 GPU. For each method, the same $512\times 512$ image is adopted as the input of ID plug-in network, and we generate a $1024\times 1024$ personalized image using the DPM-2M Karras sampler with $16$ denosing steps. We repeat the same inference process four times and record the average inference time in Table.~\ref{tab:speed}. The complete inference process consists of two parts: one is the preprocess and another is the real inference of base model. In the preprocess stage, we crop the input image, obtaining ID embedding and predicting the ID tokens (for HyperLoRA, generating the LoRA weights and merging into base model). The sequence length of LoRA coefficients is significantly larger than that of ID tokens. Thus, HyperLoRA is usually more time-consuming at this stage. However, HyperLoRA shows remarkable superiority in the second stage, since it doesn't introduce extra attention modules compared with adapter-based methods. Once merged into the base model, our HyperLoRA maintains the same inference performance as the original model.

\begin{table}[h]
    \centering
    \caption{Inference speed of different methods (milliseconds).}
    \setlength{\tabcolsep}{3pt}
    \begin{tabular}{lcccc}
        \hline
        Method & IP-Adapter & InstantID & PuLID & HyperLoRA \\
        \hline
        Preprocess & 2996 & 758 & 236 & 1143 \\
        Inference & 6148 & 8037 & 6616 & 4327 \\
        \hline
    \end{tabular}
    \label{tab:speed}
    \vspace{-5pt}
\end{table}

\subsection{Interpolation}
\begin{figure}[t]
	\centering
	\includegraphics[width=0.93\columnwidth]{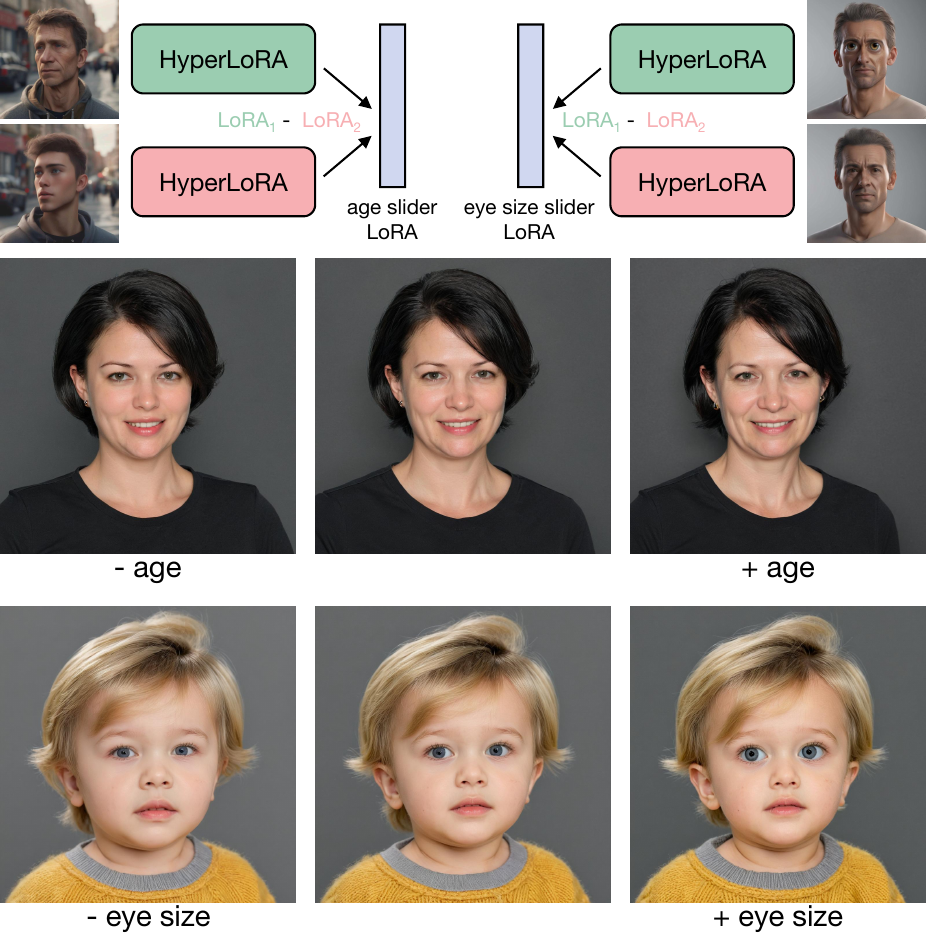}
	\caption{Using the slider LoRA generated by HyperLoRA from paired images, we are allowed to edit the local attrubutes (e.g. age or eye size) of synthesized portraits.}
	\label{fig:slider}
\end{figure}

\begin{figure*}[t]
	\centering
	\includegraphics[width=0.975\linewidth]{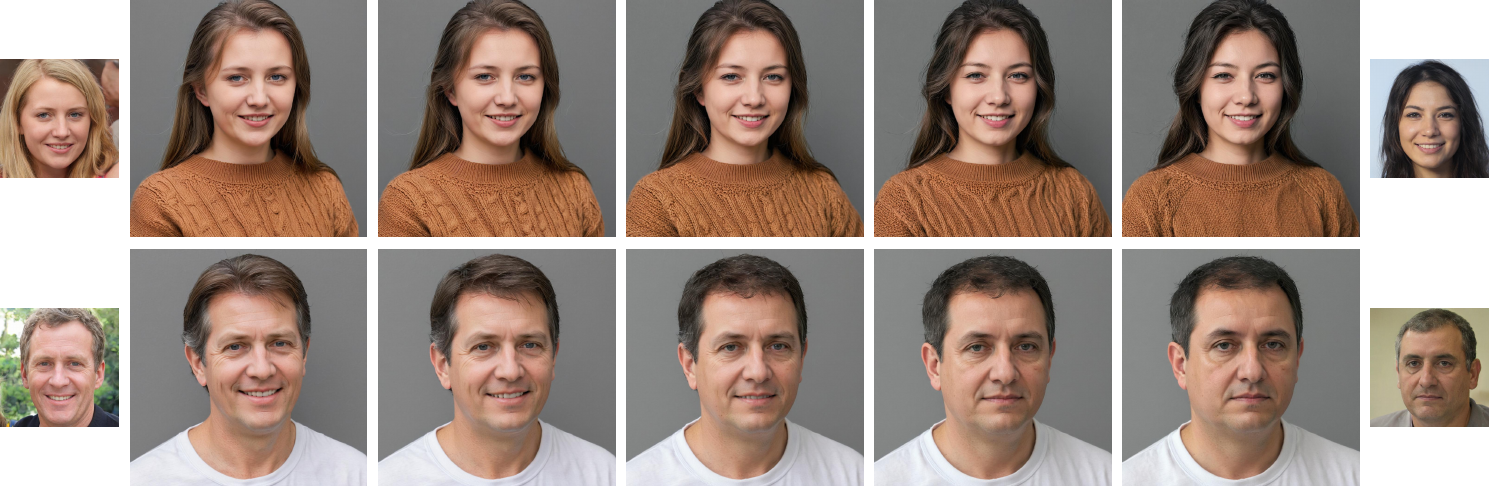}
	\caption{ID interpolation with HyperLoRA.}
	\label{fig:interp}
\end{figure*}

Benefit from the natural interpolability of LoRA, our HyperLoRA not only supports multiple input images easily but also enables smooth ID interpolation. Fig.~\ref{fig:interp} illustrates that the personalized images generated by the mixture of two Hyper ID-LoRAs can interpolate naturally from one ID to another. Furthermore, we find that the space of our generated LoRAs has similar properties as the $\mathcal{W+}$ space of StyleGAN~\cite{karras2019style}. Specifically, the slider LoRA~\cite{gandikota2024concept} can be generated by HyperLoRA given only a pair of images, as shown in Fig.~\ref{fig:slider}. The image pair consists of a normal face image and an edited one (modifying the attributes of face, e.g., age or eye size). By feeding this pair into HyperLoRA, we obtain two LoRAs, and then subtract the LoRA weights of the original image from the LoRA weights of the edited image. Fig.~\ref{fig:slider} demonstrates the difference weights between two LoRAs perform a similar behavior as a slider LoRA, possessing the ability to edit local attributes.

\subsection{More Visual Results}
\begin{figure*}[t]
    \centering
    \includegraphics[width=0.975\linewidth]{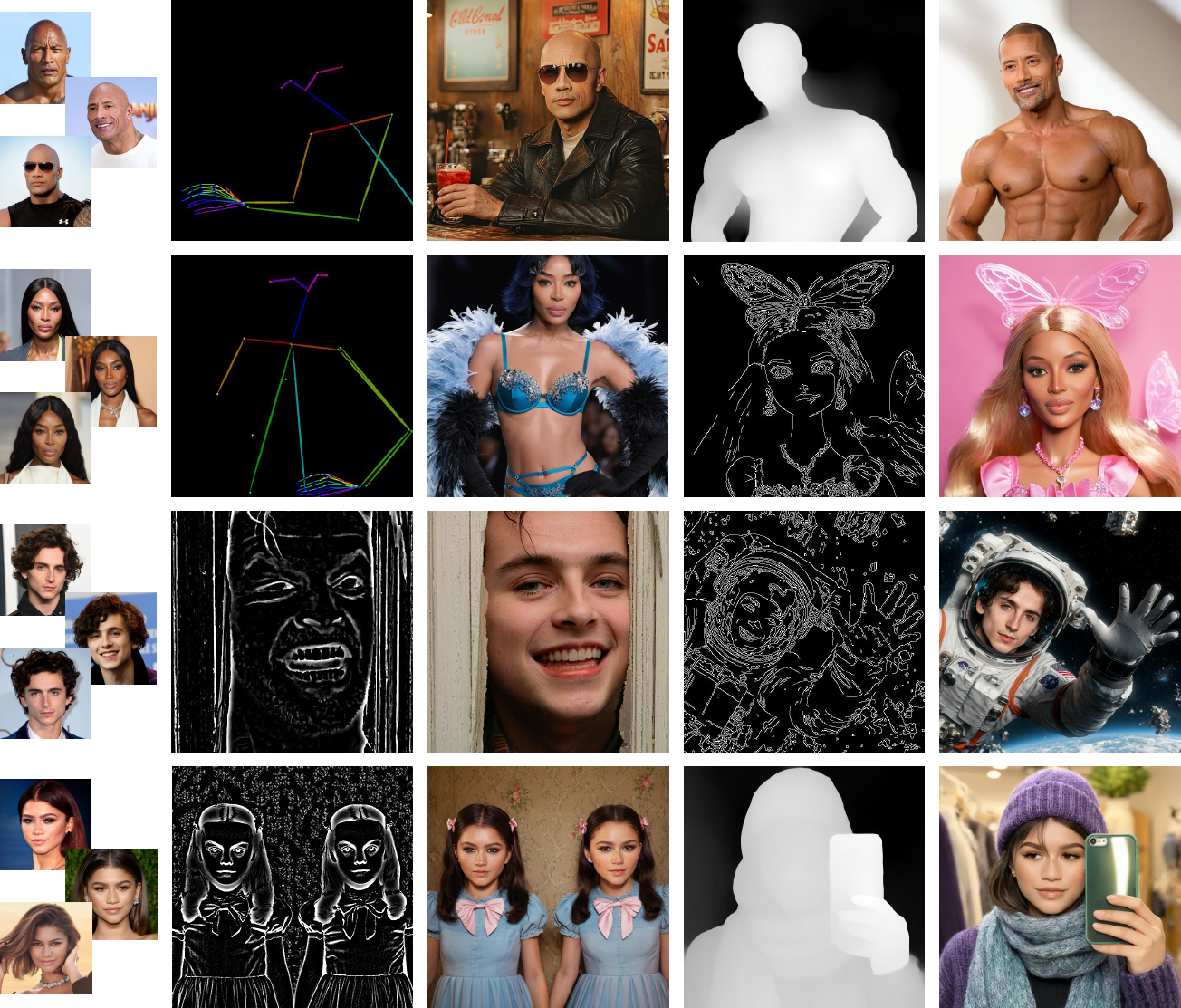}
    \caption{Portrait images generated by HyperLoRA and ControlNet.}
    \label{fig:ctrlnet}
\end{figure*}

\noindent \textbf{Inference with ControlNet.}
ControlNet~\cite{zhang2023adding} is a practical network to control the diffusion models with additional condition images (edge, pose, depth and more). It helps us to generate satisfied images more easily, getting rid of spending effort trying different seeds, prompts and other parameters. Thus, it is crucial for an ID plug-in to be compatible with ControlNet. As shown in Fig.~\ref{fig:ctrlnet}, our HyperLoRA can also generate portrait images with high photorealism and fidelity, when equipped with different ControlNets.

\begin{figure*}
    \centering
    \includegraphics[width=0.98\linewidth]{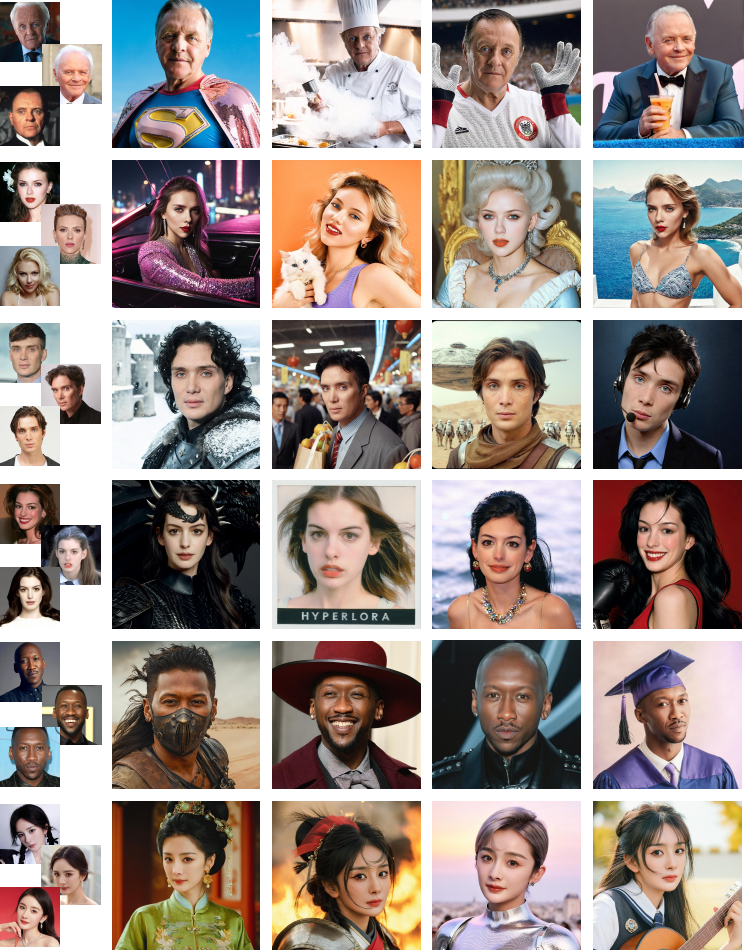}
    \caption{More Text-to-image results with HyperLoRA.}
    \label{fig:t2i_1}
\end{figure*}

\begin{figure*}
    \centering
    \includegraphics[width=0.98\linewidth]{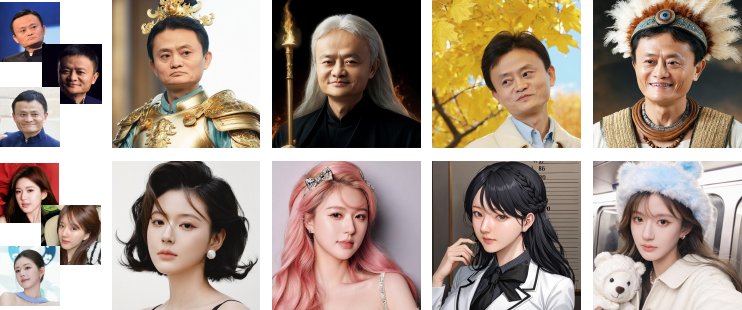}
    \caption{More Text-to-image results with HyperLoRA.}
    \label{fig:t2i_2}
\end{figure*}

\noindent \textbf{More Text-to-image Results.}
Fig.~\ref{fig:t2i_1} and Fig.~\ref{fig:t2i_2} present more portrait images synthesized by our HyperLoRA across various IDs and scenes, demonstrating its generalization.

\subsection{Discussion on Fidelity and Editability}
HyperLoRA relies on a large dataset with various facial features for training the LoRA basis, constructing a LoRA space covering sufficient face IDs, so that an unseen face can be reconstructed by the trained LoRA basis. Currently, our dataset comprises only $4.4$ million images (in contrast, InstantID is trained with $60$ million data). It is crucial to expand the dataset, aiming to further improve the fidelity. In the aspect of editability, it is a feasible solution to introduce the alignment loss of PuLID into HyperLoRA training, weakening the intrusion into the base model.

\subsection{Details of Data Filtering}
Our data filtering pipeline consists of four steps: \textbf{1)} We eliminate the images with a resolution lower than $768\times 768$; \textbf{2)} LAION Aesthetics Predictor~\cite{laionaes} is utilized to assess the aesthetics score of images, and only those with a score higher than $5.5$ are retained; \textbf{3)} We detect faces from images, and only keep those that contain a single face and the ratio of face region to the entire image is greater than $0.03$; \textbf{4)} We further filter out those images where plural personal pronouns appear in the corresponding captions.

{
    \small
    \bibliographystyle{ieeenat_fullname}
    \bibliography{main}
}

\end{document}